# Multi-objective Semi-supervised Clustering for Finding Predictive Clusters

**Zahra Ghasemi[1]\* . Hadi Akbarzadeh Khorshidi[2] . Uwe Aickelin[3]**

**Abstract**
This study concentrates on clustering problems and aims to find compact clusters that are informative regarding the outcome variable. The main goal is partitioning data points so that observations in each cluster are similar and the outcome variable can be predicated using these clusters simultaneously. We model this semi-supervised clustering problem as a multi-objective optimization problem with considering deviation of data points in clusters and prediction error of the outcome variable as two objective functions to be minimized. For finding optimal clustering solutions, we employ a non-dominated sorting genetic algorithm II approach and local regression is applied as prediction method for the output variable. For comparing the performance of the proposed model, we compute seven models using five real-world data sets. Furthermore, we investigate the impact of using local regression for predicting the outcome variable in all models, and examine the performance of the multi-objective models compared to single-objective models.

# 1 Introduction

\* Corresponding author
E-mail ad*dress: zahra.ghasemi@adelaide.edu.au*

*1 School of computer science, The university of Adelaide, Adelaide, Australia.*

*2 School of Computing and Information Systems, The University of Melbourne, Melbourne, Australia.*

*3 School of Computing and Information Systems, The University of Melbourne, Melbourne, Australia.*

Cluster analysis methods seek to partition a data set into different groups in such a way that data points belonging to each group are similar to each other. Conventional clustering methods are unsupervised, meaning that neither knowledge on partitioning of data points nor outcome variables are available (Bair, 2013).

To take advantage of both supervised and unsupervised information in clustering, semi-supervised clustering methods have emerged and attracted much interest (Alok, Saha, & Ekbal, 2015; Basu, Banerjee, & Mooney, 2002; Handl & Knowles, 2006; Saha, Ekbal, & Alok, 2012; Veras, Aires, & Britto, 2018; Yang, Sun, & Wu, 2019; Zhong, 2006). These methods can be categorized according to the nature of the known supervised information (Bair, 2013). First is the case where the data is partially labelled (Alok, et al., 2015; Basu, et al., 2002; Handl & Knowles, 2006; Saha, et al., 2012). In other words, the labels are known for a subset of observations. Second is the case where some form of relationship among some data points is known (Basu, Banerjee, & Mooney, 2004; Wagstaff, Cardie, Rogers, & Schrödl, 2001) (must-link and cannot-link constraints). Third is the case where one seeks to identify clusters associated with a particular output variable (Bair & Tibshirani, 2004; Koestler, et al., 2010). Despite the importance of the latter, relatively few studies have been conducted in this regard (Bair, 2013).

Association with output variables can be considered in various forms. In this study, we consider this association through minimizing the prediction error of the outcome variable. In other words, this study aims to propose a semi-supervised clustering method which can partition data points into dense clusters in such a way that these clusters can accurately predict the outcome variable. Our intention is not merely employing data partitioning for a more precise prediction of the output variable, which is the main purpose of cluster-wise regression methods (Zhang, 2003). We aim to find a clustering method which partitions data points with the most similar features into the same subgroups, while at the same time being informative toward the output variable. We believe that employing this type of partitioning is helpful for finding all inherent clusters and insightful for many real applications. For more clarification, consider a company that partitions its products based on their price and advertisement cost, along with their annual sales amount, as the outcome variable. Imagine that the related information for 17 products is as shown in Fig. 1. Considering just price and advertisement cost (unsupervised clustering), leads to partitioning products into two groups (A and B). However, products in group B have very different sale amounts. When guiding the clustering process using supervised information of the outcome variable (sale amount) such that for each cluster an outcome variable can be estimated with the least possible error, two new subgroups (B1 and B2) are identified. Subgroup B2 has a noticeable variation regarding the outcome variable and should be investigated more for identifying other possible effective factors.

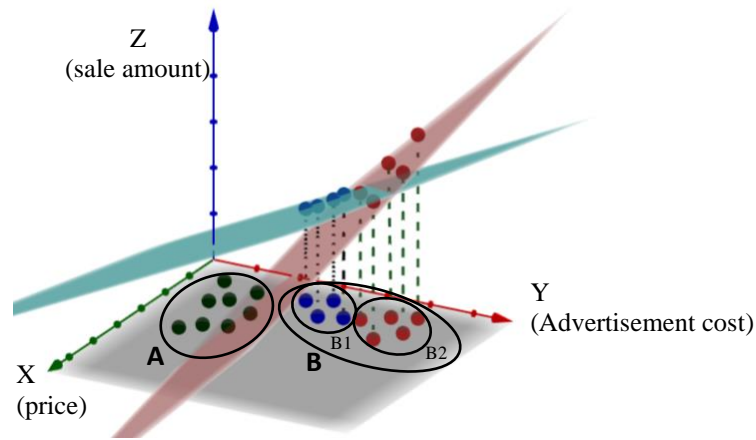

**Fig. 1** An example of semi-supervised clustering regarding an output variable

Another interesting application for semi-supervised clustering in association with the outcome variable is in healthcare for identifying all subtypes of fatal illnesses. These subtypes may not be completely recognized without considering an outcome variable, like survival time. Cancer patients, for instance, are commonly clustered based on similarity of their features into "low risk" and "high risk" groups, without considering their association with survival time (Bair, 2013). In the case of clustering patients based on similarity of their features and association with the survival time (as an outcome variable) simultaneously, patients are separated into subgroups with the most similar features and the same pattern regarding the output variable. Subgroups for which the fitted hyperplanes have a steep slope, indicate other probable factors affecting the survival time. These factors are neglected by the clustering, but we alleviate this by considering the association with the output variable. In other words, we separate subgroups with similar features, but different types of association with the outcome variable. When clustering patients considering both objective functions, all inherent subgroups of patients are recognized. Furthermore, new patients can be assigned to these more accurate clusters based on similarity of their features with patients' features in each cluster. Consequently, a more realistic estimation of their survival time is possible. This is helpful for the selection of an appropriate treatment plan for these patients.

There are few studies for semi-supervised clustering employing outcome variable values as supervised information. In some previous work, the relation with outcome variable is exploited for identifying the most important features. In other words, features are limited to those with strong effects on the output variable, and preventing irrelevant input variables to mislead the clustering process. In these studies, supervised information of output variable is used for feature selection, then common unsupervised clustering methods are applied (Bair & Tibshirani, 2004; Gaynor & Bair, 2013; Koestler, et al., 2010).

To the best of our knowledge, there is just one pervious similar study to our work. In that article, patients are clustered based on the similarity of amounts paid for medical health services in the first-week after injury in such a way that total medication cost, as the output variable, can be predicted precisely (Khorshidi, Aickelin, Haffari, & Hassani-Mahmooei, 2019). This method of partitioning can be insightful for insurance policy makers. Authors in the mentioned study modelled the semi-supervised clustering problem as a multi-objective optimization problem and employed an evolutionary approach for finding optimal solutions. Two objective functions in that study were cost function of k-median clustering and prediction error of output variable when clusters are predictors. This study is an extension of that paper. Like the previous article, we aim to find a partitioning method which can improve both objective functions. Our main contributions in this study are as follows:
- We employ local regression (LR) for output prediction.
- We employ non-dominated sorting genetic algorithm II (NSGA-II) instead of the evolutionary approach proposed in the article for improving generations.

More details about our proposed model are presented in section 3. For comparing purposes, we investigate seven other different models for the problem based on various combinations of initialization, generations improvement, and regression estimation, which is presented in detail in section 4. One of these seven models is the one developed in (Khorshidi, et al., 2019) which is discussed more in the subsequent section. Finally, we compare the performance of all eight models based on a clustering metric (deviation) and a regression measure (MAE) using five real-world data sets in section 5.

The main research objective is to identify clusters that are informative regarding the prediction of an outcome variable as well as the data points within each cluster that are similar. The research questions we intend to answer are:

RQ1. Which modelling combination (among eight models) results in the best performance in optimizing both objective functions examining using deviation and prediction error measures? We also investigate the impact of employing LR, which is the main contribution of this study.

RQ2. Can the proposed multi-objective model create clusters that are more predictive regarding the outcome variable in comparison with single-objective models?

## 2 Related Work

Semi-supervised clustering has attracted lots of research interest due to its broad applications (Li, et al., 2019). In most related research, supervised information is considered in the form of partially labelled data or known constraints (must-link and cannot-link) (Alok, et al., 2015; Basu, et al., 2002; Basu, et al., 2004; Handl & Knowles, 2006; Saha, et al., 2012; Wagstaff, et al., 2001).

In some other semi-supervised clustering studies, the aim is to identify clusters associated with an outcome variable. Despite the importance of this problem, there are relatively few researches with considering this form of supervised information (Bair, 2013). We review some of these studies in this section as this problem is the focus of our paper.

Bair et al. (Bair & Tibshirani, 2004) used the outcome variable to estimate associations between each feature and outcome variable and applied k-means clustering to the features if their association with the outcome variable were more than a user-defined threshold. In another study, a recursively partitioned mixture model (RPMM) (Houseman, et al., 2008) was used on selected features that were most strongly associated with the outcome variable (Koestler, et al., 2010). In both aforementioned studies, association with the outcome variable is used for feature selection, then traditional clustering methods are applied. RPMM, despite k-means clustering, does not require the number of clusters to be pre-determined. In another study, Gaynor and Bair (Gaynor & Bair, 2013) proposed a "supervised sparse clustering" method, which is an extension of "sparse clustering" (Witten & Tibshirani, 2010). In this proposed approach, equal initial weights are considered for features which are most strongly associated with outcome variable and these weights are updated iteratively until convergence. The main drawback of these studies is the permanent exclusion of discarded features in the initial screening, which may lead to irreversible loss of some valuable information.

In a recent study, Khorshidi et al.(Khorshidi, et al., 2019) proposed a multi-objective optimization model for clustering patients injured in transport accidents. They intended to find a method which partition patients based on both similarities of their features and their association with the total cost of medication, as the outcome variable. Two objective functions considered in this study are the cost function of k-medians clustering and the cross-validated RMSE for predicting outcome variable, when clusters are predictors. In this study input variables are amounts paid for 19 different health services in the first week after injury in transport accidents. For finding optimal solutions an evolutionary approach is presented. In the proposed approach, first, an initial pool of solutions is created with randomly selection of date points as cluster centres (RSO) and allocating other data points to these centres based on the shortest distance criterion. Solutions, then, are assessed based on two objective functions. For creating new generations, dominated solutions are improved with employing a Stochastic Gradient Descent (SGD) approach for k-medians clustering. Creating new generations continues until a maximum number of iterations is reached.

Our current paper is an extension of (Khorshidi, et al., 2019) which aims to improve both objective functions through employing different methods for updating generations, and estimating outcome variable.

## 3 Proposed Method

As stated earlier, the main aim of this study is to partition data points into different clusters such that data points in each cluster have the most similar features. For this purpose, we calculate the summation of distances between each data point and the centre of the cluster that the specific data point belongs to. Therefore, we define the first objective function in form of minimizing summation of all these distances for all data points. The second objective is the accurate prediction of the outcome variable through clustering. For this purpose, we employ a local regression method (LR) in such a way that after separating data points into clusters, a multiple linear regression equation is fitted to each cluster and the output variable is predicted using these equations. Therefore, the second objective function is defined as minimizing the prediction error of the output variable, which is the supervised section of the developed model. After defining the objectives, the problem of finding compact clusters with the ability to accurately predict outcome variables is modelled as a multi-objective optimization problem, and an NSGA-II approach is used for solving the problem. For this purpose, first, an initial pool of solutions should be created. After that, the created solutions should be assessed based on objective functions and a new population should be created with some members of the initial population and some newly created members by applying crossover and mutation as common genetic operators. Creating new populations continues until fulfilling the termination criterion, which is reaching a predetermined maximum number of iterations. More details about the model presented in this article follows.

First, we describe the objective functions. Then, we review the NSGA-II approach and its requisites (genetic representation, initialization, and genetic operators). Finally, we discuss SGD which is the improvement method employed in (Khorshidi, et al., 2019). This method is used as another approach for creating new generations and its performance is compared in section 5.

### 3.1 Objective functions

As discussed earlier, we aim to find compact clusters which are strongly associated with the outcome variable. For having compact clusters, we define first objective function as overall deviation of each clustering which is computed by Eq. 1.

$$\text{Dev}(C) = \sum_{C_k \in C} \sum_{x \in C_k} ||x - \mu_k||_1 \qquad (1)$$

Where C is the set of all clusters, and $\boldsymbol{\mu_k}$ is the centre of cluster $\boldsymbol{C_k}$. Actually, this function is the cost function of k-medians clustering which has less sensitivity to outliers compared with k-means clustering cost function (García-Escudero, Gordaliza, Matrán, & Mayo-Iscar, 2010).

For the second objective function reflecting association with outcome variable, we consider prediction error of the outcome variable. For this purpose, after partitioning observations, a hyperplane is fitted to each cluster, and the related multiple linear regression equation is determined as follows:

$$\hat{Y}_{ci} = B_0 + B_1 X_1 + \cdots + B_n X_n \qquad (2)$$

$$for\ i = 1,..,K$$

Where $\hat{Y}_{ci} = (\hat{y}_1, \hat{y}_2, \ldots, \hat{y}_{nci})'$ is a $n_{ci} \times 1$ vector of estimated output variable for cluster i, and K is the number of clusters. Then, the mean absolute error of predicted values for the output variable is calculated as formulated in Eq. 3.

$$MAE(C) = \frac{1}{K} \sum_{i=1}^{K} \frac{\sum_{j=1}^{n_{ci}} |y_j - \hat{y}_j|}{n_{ci}} \qquad (3)$$

The main aim of this study is to partition observations in such a way that both objective functions are minimized.

### 3.2 NSGA-II

After modelling the problem of clustering observations in association with outcome variable in the form of a multi-objective optimization problem, an algorithm for solving the problem needs to be selected. In this study, we select one of the most popular multi-objective optimization algorithms which is the non-dominated sorting genetic algorithm II (NSGA-II) (Deb, Pratap, Agarwal, & Meyarivan, 2002). NSGA-II is an extension of the GA which is a simple yet efficient algorithm for solving multi objective optimization problems (Deb, et al., 2002). In this algorithm, as in other evolutionary approaches, solutions are first assigned to different fronts based on their objective function values in such a way that solutions which are not dominated by other solutions constitute the ranked first front. After removing first ranked solutions, remaining non-dominated solutions constitute the second ranked front and this procedure continues until all solutions are assigned to different fronts (Kramer, 2017). For sorting purpose of the solutions that belong to the same front, another criterion named "crowding distance", which is the Manhattan distance of two neighbouring solutions for two objectives, is computed. When applying binary tournament selection with crowding-comparison, parents are determined, and an offspring population is created through applying crossover and mutation operators. Offspring and current generation population are combined, and all solutions are sorted based on front rank and crowding distance values and the best solutions form the new generation.

When applying NSGA-II to solve the semi-supervised clustering problem, it is necessary to select an appropriate genetic representation. Furthermore, a pool of initial solutions should be created, and genetic operators need to be specified. These issues are investigated in more detail in the subsequent sections.

### 3.2.1 Genetic Representation

In this study, a locus-based adjacency representation proposed by Park et al. (Park & Song, 1998) is employed. In this genetic representation, each genotype is comprised of N genes $g_1,…,g_N$, where N is the number of observations in a data set. Each gene can take integer values from 1 to N as alleles. Assigning a value of j to the ith gene indicates that there is a link between observations i and j. For instance, as it can be seen in Fig. 2, the assigned number to the first gene is 3 which indicates an arrow starts from data point 1 and ends at data point 3. Similarly, for position 6, the related number in the genotype is 7, indicating an arrow from data point 6 to 7. Eventually, all connections are identified and those observations which are connected through these links are partitioned into the same cluster, as illustrated in Fig. 2. standard crossover and mutation operators can be applied without difficulty with this representation.

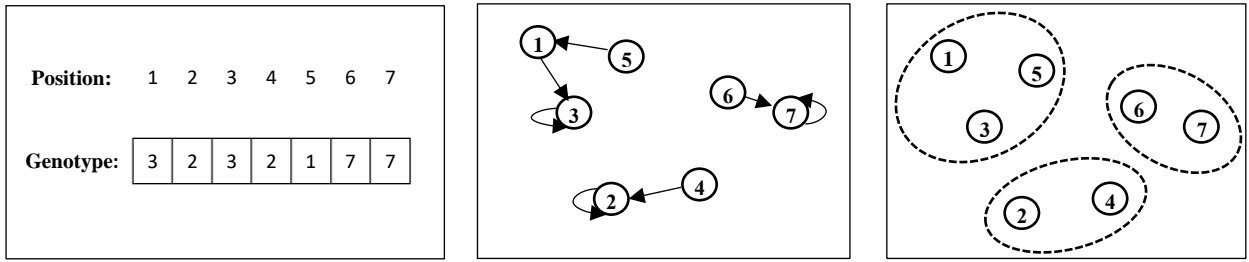

**Fig. 2** Genetic Representation

### 3.2.2 Initialization

For creating an initial pool of clustering solutions, we use the RSO approach (Khorshidi, et al., 2019). This approach is a purpose built initialization method for creating initial solutions in conformity with the deviation objective function. In this method, k data points are selected randomly and the distance matrix between each data point and the selected centres is calculated. Finally, each data point is assigned to the centre with the shortest distance.

Since there is no prior information about the number of clusters, as suggested in (Khorshidi, et al., 2019), the number of clusters (k) is selected from two to ten. Thus, for creating the first clustering solution, 2 random observations are selected as the centres of the clusters. The second member of the initial population is created by selecting 3 data points as cluster centres, and this process continues until selecting 10 random data points as cluster centres. This process is repeated until generating all members of the initial population.

### 3.2.3 Genetic Operators

In this study, uniform crossover and swap mutation are applied for creating new generations (Hong, Wang, Lin, & Lee, 2002). In uniform crossover, one mask with random values of 0 and 1 is generated, and each allele of an offspring is taken from the first or second parent if the corresponding value in the mask is 0 or 1, respectively. This crossover operator is unbiased and has the capability of generating any combination of parents (Whitley, 1994). In swap mutation, two random genes are selected, and their alleles are exchanged. These operators are illustrated in Fig. 3.

| **Uniform Crossover** | | | | | | | | **Swap Mutation** | | | | | | | |
|---|---|---|---|---|---|---|---|---|---|---|---|---|---|---|---|
| **Parent 1:** | 3 | 2 | 3 | 2 | 1 | 7 | 7 | **Parent:** | 3 | 2 | 3 | 2 | 1 | 7 | 7 |
| **Parent 2:** | 4 | 1 | 3 | 7 | 5 | 1 | 6 | **Offspring** | 3 | 1 | 3 | 2 | 2 | 7 | 7 |
| **Mask** | 0 | 1 | 0 | 0 | 1 | 1 | 0 | | | | | | | | |
| **Offspring** | 3 | 1 | 3 | 2 | 5 | 1 | 7 | | | | | | | | |

**Fig. 3** Crossover and mutation operators

### 3.3 Stochastic Gradient Descent

The updating method for creating new generations in (Khorshidi, et al., 2019) is SGD for k-medians clustering (Cardot, Cénac, & Monnez, 2012). In this method, clustering solutions are improved through updating cluster centres using Eq. 4.

$$X_{n+1}^r = X_n^r - a_h^r I_r(Z_n; X_n) \frac{X_n^r - Z_n}{||X_n^r - Z_n||} \qquad (4)$$

Where $X_n^r$ is the centre of the rth cluster in nth iteration. $Z_n$ is a randomly selected observations in the selected cluster to improve, and $a_h^r$ is the learning rate computed as:

$$a_n^r = \frac{c_\gamma}{(1+c_\alpha n_r)^\alpha} \qquad (5)$$

Where $n_r$ is the number of observations in the rth cluster and α, cγ, and cα are constant parameters. Dominated solutions in each generation are improved using Eq. 4. Parameters' values for the SGD method are set based on the proposed amounts in (Khorshidi, et al., 2019) as presented in Table 1.

### 4 Experimental design

In this section, eight models and their settings are discussed. Implementation results of models are presented in the subsequent section. In our model for this study, we employ RSO for initialization, LR as the predicting method of the output variable, and crossover and mutation (CM) as the updating method in each generation.
In order to compare results obtained with the proposed model, we also employ the methods used in (Khorshidi, et al., 2019) for output prediction and updating generations. For predicting the outcome variable, clusters are considered as predictors (CP). The employed regression equation is as follows:

$$Y = B_0 + B_1 X \qquad (6)$$

where X is the vector of cluster labels for all data points. SGD is the improvement method of generations applied in (Khorshidi, et al., 2019) as described in section 3.3. Furthermore, we examine another initialization method which is random chromosomes (RC) for creating completely random initial clusters. In this method, for each clustering solution to partition N data points, one random chromosome with N genes is created, and for each gene one random integer number from 1 to N is selected as allele. Applying this initialization method have two major advantages. Firstly, no prior information about the number of clusters is required. Furthermore, all possible solutions have the equal chance for being selected.

An overall overview of different models is presented in Fig. 4. Models' parameters are set based on values in Table 1. It is worth noting that parameter values for the SGD method are selected based on amounts proposed in (Khorshidi, et al., 2019) and crossover and mutation percentages are selected based on values suggested in (Hassanat, et al., 2019), as commonly used values in literature. When considering different initialization, regression estimation and updating generations methods, eight models for clustering data points in association with outcome variable are developed as described in Table 2. Except for model 5, which is the model developed in (Khorshidi, et al., 2019), all other seven models in Table 2 are developed in this study for the first time. For comparing results of the eight models, we use five data sets as described in Table 3.

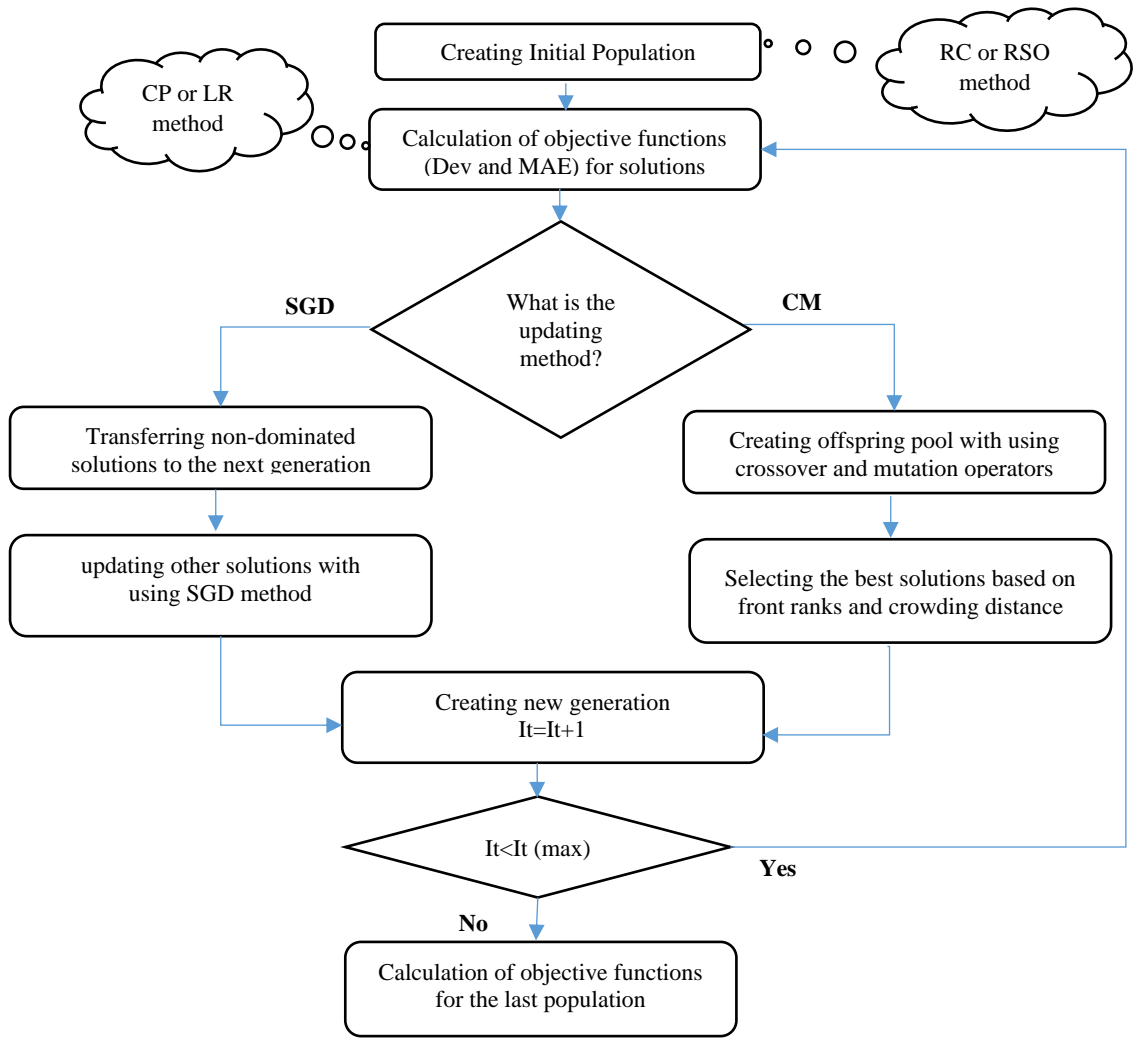

**Fig. 4** Overall overview of different models

**Table 1** Parameters settings

| | | |
|---|---|---|
| General Parameters | Population size<br>Number of Iteration | 100<br>100 |
| CM parameters | Crossover type<br>Mutation type<br>Crossover Percentage<br>Mutation Percentage | Uniform<br>Swap<br>90<br>3 |
| SGD parameters | $c_\gamma$<br>$c_\alpha$<br>$\alpha$ | 2000<br>1<br>0.75 |

**Table 2** Developed models

| Model No. | 1 | 2 | 3 | 4 | 5 | 6 | 7 | 8 |
|---|---|---|---|---|---|---|---|---|
| Initialization | RSO | RC | RSO | RC | RSO | RC | RSO | RC |
| Regression method | CP | CP | LR | LR | CP | CP | LR | LR |
| Updating method of generations | CM | CM | CM | CM | SGD | SGD | SGD | SGD |

**Table 3** Specification of data sets

| Data Set No. | Name | Number of Attributes | Number of Instances | Link |
|---|---|---|---|---|
| 1 | Boston Housing | 14 | 506 | https://www.cs.toronto.edu/~delve/data/boston/bostonDetail.html |
| 2 | Concrete Compressive Strength | 9 | 1030 | https://archive.ics.uci.edu/ml/datasets/Concrete+Compressive+Strength |
| 3 | QSAR Aquatic Toxicity | 9 | 546 | https://archive.ics.uci.edu/ml/datasets/QSAR+aquatic+toxicity |
| 4 | Airfoil Self-Noise | 6 | 1503 | https://archive.ics.uci.edu/ml/datasets/Airfoil+Self-Noise |
| 5 | Concrete Slump Test | 10 | 103 | https://archive.ics.uci.edu/ml/datasets/Concrete+Slump+Test |

## 5 Results

Implementation results for each of the developed models using different datasets are obtained. Fig. 5 shows one sample of implementation results in form of the average objective function values in different iterations.

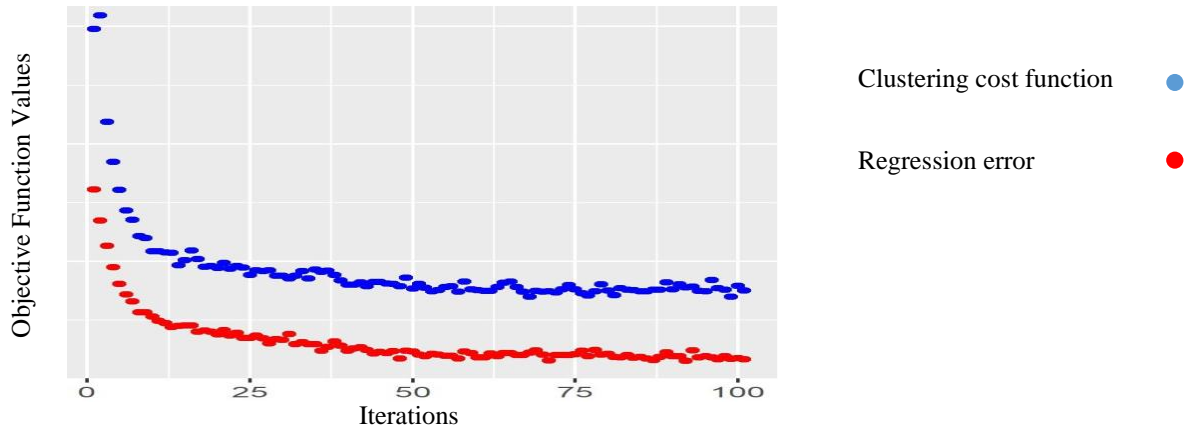

**Fig. 5** Average objective function values in different iterations for model no. 5 using dataset no. 2

For determining the best performing models, statistical tests are performed on results obtained using each of five datasets. In the following, as an example, the comparison method for determining the best performing models in term of deviation objective function using data set no. 2 is presented in detail. A Box Plot of the deviation objective function of the last generation for different models using data set no. 2 is illustrated in Fig. 6.

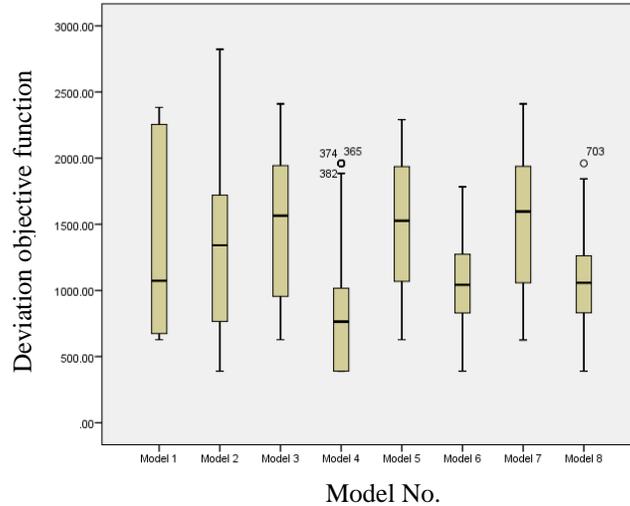

**Fig. 6** Box Plot of deviation objective function for different models using data set no. 2

For determining the best performing model, the values of the deviation objective function in the last generation for 8 models are compared. When considering 100 objective function values in the last generation for each model, it is assumed that these values are normally distributed and the one-way analysis of variance (ANOVA) is used to determine whether there are statistically significant differences among the means of deviation objective function values for different models. If the calculated p-value is less than 5% significance level, we confirm a significant difference among the results of the models. For determining the best performing models, the model with the least mean value for deviation objective function is chosen and Tukey's test is applied to identify other models where their difference with the selected model is not statistically significant. Results of Tukey's test are presented in Table 4. Table 5 shows different subsets of models for alpha = 0.05.

**Table 4** Tukey test results for deviation values of eight models using data set number 2

| Dependent Variable | MODEL | Sig. | | | | | | | |
|---|---|---|---|---|---|---|---|---|---|
| | | Model 1 | Model 2 | Model 3 | Model 4 | Model 5 | Model 6 | Model 7 | Model 8 |
| Deviation | Model 1 | | 1.000 | .985 | .000 | .918 | .000 | .897 | .000 |
| | Model 2 | 1.000 | | .950 | .000 | .823 | .000 | .791 | .001 |
| | Model 3 | .985 | .950 | | .000 | 1.000 | .000 | 1.000 | .000 |
| | Model 4 | .000 | .000 | .000 | | .000 | .892 | .000 | .798 |
| | Model 5 | .918 | .823 | 1.000 | .000 | | .000 | 1.000 | .000 |
| | Model 6 | .000 | .000 | .000 | .892 | .000 | | .000 | 1.000 |
| | Model 7 | .897 | .791 | 1.000 | .000 | 1.000 | .000 | | .000 |
| | Model 8 | .000 | .001 | .000 | .798 | .000 | 1.000 | .000 | |

According to the obtained results, model number 4 results in the least mean of deviation errors, and the difference between results of this model with models number 6 and 8 are not statistically significant. Consequently, models number 4, 6 and 8 are the best performing models for the deviation objective function using data set number 2. Similar tests have been performed for other objective function and other data sets and results are shown in Table 6.

As it is evident in Table 6, models 3,7 and 8 result in the best values for both objective functions while using data set number 3. In all these models, LR is employed for estimating output variable. Updating method of generations for models 7,8 is SGD and for model 3 is CM. Initialization method employed in the mentioned models are RSO for models 3, 7 and RC for model 8.

**Table 5** Subsets for alpha = 0.05 for deviation values of eight models using data set number 2

| MODEL No. | Number of data points | Subset for alpha = 0.05 | |
|---|---|---|---|
| | | 1 | 2 |
| Model 4 | 100 | 915.1014 | |
| Model 6 | 100 | 1019.1698 | |
| Model 8 | 100 | 1034.7571 | |
| Model 2 | 100 | | 1371.4854 |
| Model 1 | 100 | | 1389.3287 |
| Model 3 | 100 | | 1460.8188 |
| Model 5 | 100 | | 1487.5328 |
| Model 7 | 100 | | 1492.2259 |
| Sig. | | .798 | .791 |

**Table 6** The best performing models for different data sets

| | Best models for each dataset | | | | |
|---|---|---|---|---|---|
| | DS No. 1 | DS No. 2 | DS No. 3 | DS No. 4 | DS No. 5 |
| Both Measures | - | - | 3,7,8 | - | - |
| Clustering Measures (Dev) | 6,8,1,5 | 4,6,8 | 1,3,5,6,7,8 | 1,5,6,8 | 2,4,6,8 |
| Regression Measures (MAE) | 3 | 3 | 3,4,7,8 | 3 | 3 |

Model 3, which is the main model developed in this study with RSO as initialization method, LR as prediction method of outcome variable and CM as updating method of generations, results in the least values for regression objective function for all data sets and the best values of deviation objective function for one of data sets. Thus, it can be concluded that model 3 outperforms others in comparison with its 7 counterparts. Furthermore, in all models with the best results for the MAE objective function (models 3,4,7,8) LR is applied for output estimation which proves the positive effect of using LR for output prediction. Consequently, the answer of our first research question is that the best performing model which minimize both objectives simultaneously is model 3 and employing LR for outcome prediction is most efficient.

For finding the answer of our second research question, which is whether using multi-objective approach results in finding clustering solutions which are more informative regarding outcome prediction, we developed a single-objective models with just the deviation objective function and compared obtained results with multi-objective models. For this purpose, in models 1 to 4, instead of using NSGA-II, we employed a conventional Genetic Algorithm (GA) with deviation objective function, and compared MAE values in the last generation using Independent Samples t Test. This statistical test is used when two independent groups are compared in terms of their mean values. One sample of comparing results using data set no. 4 is shown in Table 7.

As it is evident in this table, in all four cases, multi-objective models result in lower values for the regression measure (MAE) and these differences are statistically meaningful according to our significance level. Similar results are obtained when other data sets are used, which confirms the effectiveness of multi-objective modelling for finding clusters which can predict outcome variable more accurately.

**Table 7** Comparison results for multi-objective and single-objective models using data set no. 4

|  | Model 1 | | Model 2 | | Model 3 | | Model 4 | |
| --- | --- | --- | --- | --- | --- | --- | --- | --- |
|  | Multi-objective | Single-objective | Multi-objective | Single-objective | Multi-objective | Single-objective | Multi-objective | Single-objective |
| Average MAE | 6.24 | 14.02 | 6.87 | 26.09 | 3.213128 | 4.799244 | 3.785270 | 4.803110 |
| Sig. | 0.000 | | 0.000 | | 0.000 | | 0.000 | |

## 6 conclusions

In this study, the problem of clustering data points with the aim of identifying dense clusters which are meaningfully associated with outcome variable is modelled as a multi-objective optimization problem with considering deviation of data points and prediction error of outcome variable as objective functions to be minimized. In the main proposed model (model 3), RSO is the chosen initialization method which is a purposeful initialization method that finds initial solutions aligned with deviation objective function. For predicting outcome variable in this model, LR is used which employs more information from input variables for prediction purpose, compared with CP method. Updating method of generations in this model is CM which tries to improve solutions in terms of both objective functions. For comparison purposes, we developed 6 other models through combining methods proposed in (Khorshidi, et al., 2019) for initialization (RSO), generations improvement (SGD), and output estimation (CP) with RC initialization (Park & Song, 1998), CM, and LR. Although all these methods, except for LR, have already been introduced in previous studies, they have been combined here for the first time to solve the concerned problem. For identifying the best performing model, we compared the performance of the main developed model with seven counterparts, one of which is the proposed model in (Khorshidi, et al., 2019), based on clustering and regression measures using five real-world data sets (addressing RQ1). Comparison results showed that the main developed model (model 3) is the best performing one among all eight models. Furthermore, employing LR resulted in more accurate prediction of outcome variable. Moreover, obtained results confirm that using multi-objective approach results in less prediction error for outcome variable compared to single-objective models (addressing RQ2).

The proposed approach can be applied to identify clusters of data points where their members have the most similar features and the related output variable can be predicted with the least possible error simultaneously. One interesting area of application for the proposed model is in healthcare for clustering patients with fatal diseases, when survival time is considered as the outcome variable. When applying the model proposed in this study, it is possible to separate patients into distinct clusters in such a way that those patients which are assigned to the same cluster have the most similar features and identified clusters have the same pattern regarding the output variable. This approach is beneficial for uncovering complex disease subtypes which probably may not be detected accurately with considering just input features. Having this precise clustering, new patients can be assigned to the appropriate cluster based on similarity of their features with patients' features belonging to identified clusters and it is possible to predict their survival time with the least possible error, and an appropriate medical plan can be selected for them.

For future research, it is possible to enhance the proposed model to make it more robust for tackling uncertain real datasets. Furthermore, one may work on enhancing initialization and generations improvement methods to be aligned with both objective functions.

**Declarations**
**Funding** This research did not receive any specific grant from funding agencies in the public, commercial, or not-for-profit sectors.
**Conflicts of interest** The authors declare that they have no conflict of interest.